%% file: main.tex
\documentclass[acmtog, nonacm]{acmart}
\acmSubmissionID{1812}
\AtBeginDocument{%
  }

\usepackage{makecell}
\usepackage{color, colortbl}

\newcommand{\methodname}{{VDAWorld}}

\newcommand{\rebuttal}[1]{\textcolor{black}{#1}}
\newcommand{\sig}[1]{\textcolor{black}{#1}}

\colorlet{bestcolor}{blue!30}
\colorlet{secondbestcolor}{blue!15}
\colorlet{thirdbestcolor}{gray!15}
\newcommand{\best}[1]{\cellcolor{bestcolor}#1}
\newcommand{\second}[1]{\cellcolor{secondbestcolor}#1}

\usepackage{wrapfig}
\begin{document}

\title{VDAWorld: World Modelling via VLM-Directed Abstraction~and~Simulation}

\author{Felix O'Mahony}
\email{foo222cam.ac.uk}
\author{Roberto Cipolla}
\author{Ayush Tewari}
\affiliation{%
  \institution{University of Cambridge}
  \country{United Kingdom}
}

\renewcommand{\shortauthors}{O'Mahony et al.}

\begin{abstract}
  Generative video models, a leading approach to world modeling, face fundamental limitations. They often violate physical and logical rules, lack interactivity, and operate as opaque black boxes ill-suited for building structured, queryable worlds. 
  To overcome these challenges, we propose a new paradigm focused on distilling an image caption pair into a tractable, abstract representation optimized for simulation. We introduce \methodname{}, a framework where a Vision-Language Model (VLM) acts as an intelligent agent to orchestrate this process. 
  The VLM autonomously constructs a grounded (2D or 3D) scene representation by selecting from a suite of vision tools, and accordingly chooses a compatible physics simulator (e.g., rigid body, fluid) to act upon it. 
  \methodname{} can then infer latent dynamics from the static scene to predict plausible future states. Our experiments show that this combination of intelligent abstraction and adaptive simulation results in a versatile world model capable of producing high quality simulations across a wide range of scenarios.
  \sig{We demonstrate several applications of \methodname{} across interactive control and counterfactual generation, and physical and logical reasoning, achieving state-of-the-art results on several benchmarks.}
\end{abstract}

\begin{teaserfigure}
  \centering
  \includegraphics[width=0.9\textwidth]{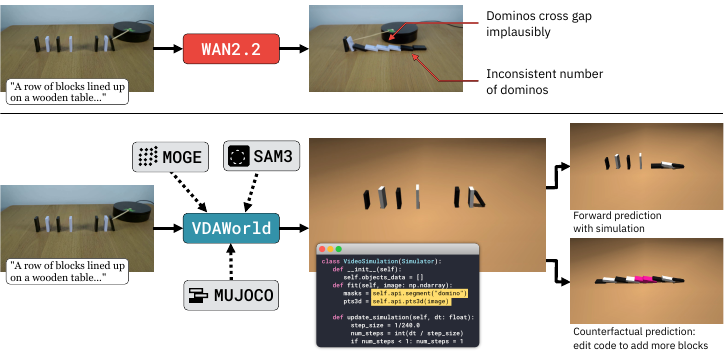}
  \caption{Given an image and caption pair, \methodname{} builds a complete Pythonic simulator of the scene, resulting in an interactive, interpretable world model}
  \Description{}
  \label{fig:teaser}
\end{teaserfigure}

\maketitle
\input{sections/intro}

\input{sections/related}
\input{sections/method}

\input{sections/results}

\input{sections/limitations.tex}
\input{sections/conclusion}
\input{sections/acknowledgements}
\bibliographystyle{ACM-Reference-Format}
\bibliography{ref}

\include{sections/appendix}

\end{document}

%% file: sections/intro.tex
\section{Introduction}
Understanding and forecasting how the visual world evolves is a core challenge for building intelligent systems. 
Humans can observe a static scene and infer not only its current structure but also how it might change over time and in response to different actions. 
This capability underlies essential skills such as planning, decision-making, and causal reasoning.
A central hypothesis for how humans perform this reasoning is ``world modelling'', formalised as the process of constructing an internal, mental representation of the external world that can be used to predict future states and reason about cause and effect~\citep{craik_1944_world_models}.
Replicating this physical intuition in AI hinges on creating equivalent models to predict potential futures. 

In recent years, the dominant paradigm for building such models has been large-scale video generation, e.g., Genie~\citep{bruce_2024_genie}. 
By training on immense visual corpora, these models have achieved remarkable success in synthesizing complex, dynamic scenes with impressive visual realism, suggesting they are learning a powerful, albeit implicit, model of our world only from 2D observations and text descriptions~\citep{wan2025}.
However, despite their visual prowess, pixel-space models result in critical and systematic failures that limit their use as world models. 
First, they frequently generate physically implausible scenarios~\citep{physicsiq,phyworld}. 
Video models learn statistical correlations from pixels and do not enforce any physical plausibility constraints, which can result in their outputs often violating fundamental principles of object permanence, collision, and causality.  
This failure to deduce underlying principles is not limited to 3D physics; we show that these models are equally unable to infer the structured representations and simple, deterministic rules required to simulate abstract 2D environments like Conway's Game of Life~\citep{conway1970game} (Figure \ref{fig:conwayBench}).
Second, these models operate as opaque ``black boxes''.
The generated scene is not a structured, queryable world but a sequence of pixels. 
Consequently, it is impossible to inspect the environment's underlying state or apply novel physical actions beyond those observed during training.  
Finally, as most video models are only conditioned on image and text inputs, they lack a native mechanism for physical interaction. %
While some models incorporate action conditioning~\citep{song2025history,genie3, wang2023drivedreamer}, the action space is typically limited to simple transformations, such as camera viewpoint, rather than complex physical interventions, and
where models do incorporate physical interventions, these are typically only for narrow application domains such as  driving~\citep{wang2023drivedreamer}.

Separate from video models, another significant line of research has focused on reconstructing 3D or 4D scene representations from images. 
Foundational models like DUST3R~\citep{dust3r} have demonstrated impressive capabilities in producing dense and accurate geometric reconstructions, while methods based on Neural Radiance Fields (NeRFs)~\citep{mildenhall2021nerf} excel at generating photorealistic novel views. 
However, the primary objective of both lines of work is to capture a scene’s geometry and appearance, not its underlying physical nature. 
This focus on 3D geometry also renders them, by design, inapplicable to abstract 2D environments governed by logical rules, such as cellular automata like Conway's Game of Life.
PhysGen3D~\citep{chen2025physgen3d} and PhysGen~\citep{physgen} reconstruct scenes as simulations within fixed simulators.
However, because these methods use a rigid pipeline, they are not applicable in a wide range of settings, and are exclusively able to simulate 3D material point and 2D rigid body scenes respectively.

In this paper, we introduce \methodname{}, a novel framework for world modeling that moves away from direct pixel prediction and instead builds an explicit, structured world representation. 
Instead of predicting pixels, our primary goal is to distil a visually complex image into a tractable abstract representation (Figure \ref{fig:teaser}). 
This representation intentionally discards physically irrelevant information (like fine-grained textures or static backgrounds) to create a structured world model optimized for simulation.
The use of simulators enforces structural plausibility, and provides an easy framework for intervention and action inputs within the model.

Our framework achieves this through a Vision-Language Model (VLM) that acts as a central agent, orchestrating three core innovations (see Figure~\ref{fig:method}). First, the VLM
acts as an intelligent tool-using agent to construct a grounded representation. 
It is equipped with a versatile suite of vision modules, including segmentation, 3D reconstruction, and 3D mesh and primitive fitting, and autonomously decides which tools to deploy. 
This allows the representation to be grounded in the scene’s native dimensionality; for instance, applying a full 3D pipeline for spatial environments while recognizing that such tools are irrelevant for planar ones, such as Conway's Game of Life.
Second, the choice of representation and simulator is co-dependent and adaptive. 
The VLM jointly determines the type of abstraction and the most appropriate physics simulator to act upon it: a scene with blocks is abstracted into a rigid body model and paired with a rigid body solver, while one with water is represented as a particle system paired with a fluid dynamics engine.
Finally, this structured world model enables \methodname{} to infer latent dynamics, predicting a scene's likely evolution from the image and caption alone based on visual cues and the description of the scene. \

Through comprehensive experiments, we show this combination of intelligent abstraction, adaptive simulation, and inferred dynamics results in a world model that significantly outperforms prior methods in producing high-quality, physically and logically plausible simulations across a wide range of scenarios.
\sig{
Unlike video models, the output of our method is natively controllable, amenable to user fine-grained interactions that are far beyond training data, e.g., updating the gravity in a scene. 
}

In summary, our contributions are as follows:
    (i) A new paradigm for world modelling that builds general purpose world models through structured simulation-ready representations, rather than pixel prediction, 
    (ii) a framework where a VLM acts as an intelligent agent and an inverse dynamics engine to construct a grounded scene representation from an image-caption pair, leveraging a diverse toolbox of state-of-the-art computer vision tools, and 
    (iii) demonstration of key advantages of this approach, including  physical plausibility, general interactivity, and simulation times which naturally scale with the complexity of the underlying scene dynamics. Alongside evaluations on two existing benchmarks, we propose two additional benchmarks to evaluate physical and logical reasoning.

%% file: sections/related.tex
\section{Previous Work}
\paragraph{Representation Learning with Videos}
Recent advances in generative methods have established large-scale video models as a dominant paradigm for modeling world dynamics. 
State-of-the-art models have demonstrated a remarkable ability to synthesize high-fidelity and temporally coherent videos from text and image inputs~\citep{sora,veo3,gen4,wan2025}. 
These models use diffusion or flow-matching approaches in a compressed latent space for video generation. 
While most models only condition on image and text inputs, some recent methods have enabled conditioning on other parameters, such as camera parameters, for more controllable video generation~\citep{huang2025voyager,song2025history,genie3}.

Additionally, some models demonstrate conditioning on action inputs. Many of these tend to be limited to action in a controlled domain, such as robotic manipulation tasks~\citep{fangqi_irasim}, egocentric pose~\citep{bai_2025_whole_body} or driving~\citep{wang2023drivedreamer}.
More general approaches for action conditioning include the Dreamer architecture~\citep{hafner_2025_dreamer} or DreamDojo~\citep{gao_2026_dreamdojo}.
However, it is very difficult to interact with these models outside of the specific control parameters developed during training. 
It is generally impossible to query the state of an object, apply a novel physical force, or explore alternative outcomes under different conditions. 
These models do not explicitly reason in a structured space, and instead directly perform computations in frame space. 
This leads their outputs to frequently violate fundamental principles of the real world~\citep{physicsiq,pisa,phyworld}.
Common failure cases include the violation of object permanence, where objects may inexplicably appear or vanish, and inconsistent causality, where actions do not have plausible consequences.

Video Joint Embedding Predictive Architecture (V-JEPA)~\cite{lecun_vjepa_2024} learns to predict latent representations of the world rather than raw pixels.
Further versions, such as Action Conditioned V-JEPA 2~\cite{lecun_2025_vjepa2} additionally incorporate action conditioning, allowing the model to condition its predictions on actions and future states.
However, the latent space is still learned implicitly from data and does not guarantee physical plausibility.
Additionally, the latent representations are not necessarily interpretable or queryable, and the action conditioning is limited to specific transformations rather than general interventions.
Finally, these models still operate as black boxes, and do not provide a mechanism for user intervention or control over the generated world.

\input{figures/method}
\paragraph{Scene Reconstruction and Simulation}
3D reconstruction from images has achieved considerable success in recent years. 
Models like DUST3R~\citep{dust3r} and its follow-ups~\citep{monst3r,vggt,st4rtrack2025} produce dense geometric reconstructions from images, while methods based on Neural Radiance Fields (NeRFs)~\citep{pixelnerf,dfm} and 3D Gaussians (3DGS)~\citep{splatter,pixelsplat} and their 4D extensions~\citep{4DGS,nrnerf,yang2023deformable3dgs,yunus2024recent} excel at generating photorealistic novel views. 
While most approaches do not inherently decompose the scene into discrete, object-centric components, some methods have made progress with the help of features from vision-language models~\citep{kerr2023lerf,jatavallabhula2023conceptfusion}. 
Additionally, most neural radiance methods are optimized for appearance rather than physics, making them poorly suited for interactive simulation. 
Some work has attempted to model physics, for instance, by leveraging fixed physics engines and 3D scene representations~\citep{physdreamer,modalnerf,pienerf,pacnerf,galileo,le2025pixie,chen2025physgen3d,kairanda2025thin,xie2024physgaussian} or 2D scene representations~\citep{physgen}.
Two works similar to ours, PhysGen3D~\citep{chen2025physgen3d} and PhysGen~\citep{physgen}, reconstruct object-centric 3D and 2D scenes respectively and perform simulation with a fixed physics engine.
The use of a fixed simulator limits the generalisability of these methods across different types of scenes and dynamics, where different simulators might be more appropriate.
Additionally, the rigidity of these approaches means they are less robust to failure cases in reconstruction, which often leads to catastrophic failures in simulation.
In this work, we use tools from scene reconstruction and simulation methods but do not use a fixed pipeline as the VLM is free to select scene representations and simulators best suitable for any input.

\paragraph{Program Synthesis with VLMs}
Our work is informed by recent advances in using Vision-Language Models (VLMs) as agents that synthesize programs to solve complex visual tasks.
A foundational paradigm is visual program synthesis for querying.  Methods like ViperGPT~\citep{suris2023vipergpt} and VisProg~\citep{gupta2023visual} can parse a complex visual query into a sequence of steps, generating code that calls various vision APIs (e.g., object detectors, depth estimators) to arrive at a final answer. 
LayoutGPT~\citep{feng2023layoutgpt} uses an LLM to generate a complete scene layout, including the sizes, positions, and relationships of different objects.
Other research has shown that a VLM can evolve interpretable visual classifiers~\citep{chiquier2024evolving} and design interpretable programs to describe underlying scientific laws~\citep{mall2025disciple}.
A second major application is in high-level planning and robotics. 
This research aims to create agents that can reason about the world to perform actions. 
VisualPredicator~\citep{liang2024visualpredicator}, for example, learns neuro-symbolic predicates that classify the state of the world for a symbolic planner. 
Others, like VoxPoser~\citep{huang2023voxposer}, use LLMs to synthesize 3D affordances that guide a low-level motion planner.
Finally, some work has explored the use of VLMs to guide editing in structured 3D environments, such as BlenderAlchemy~\citep{huang_2024_blenderalchemy}.
Concurrent work, Vision as Inverse Graphics~\citep{yin_2026_viga}, also explores the use of VLMs to generate interactive scenes.
However, this approach is focused on generating a single 3D scene representation from an image, and does not explore the generation of a simulator or the inference of dynamics.
The common thread in this research is that the VLM's role is to generate a plan or a set of actions for an agent to execute within an existing environment.

%% file: figures/method.tex
\begin{figure*}[t!]
    \centering
    \includegraphics[width=506pt]{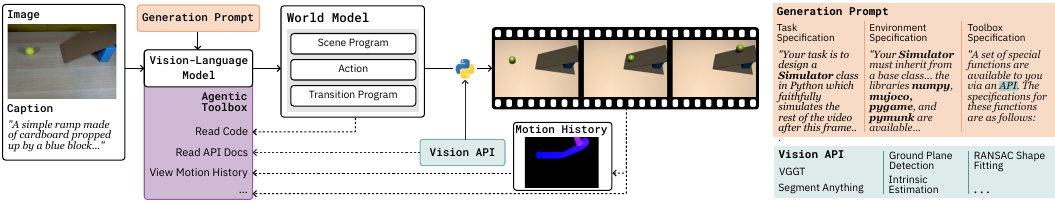}
    \caption{\rebuttal{Illustration of our method. \methodname{} takes a single image and text caption as input. %
    We design a generation prompt that the VLM uses to generate a model of the scene, including the scene program, actions, and a transition program. 
    The simulator code is then executed to generate future predictions. 
    We can also generate other diverse videos by interactively changing the actions. 
    The VLM operates in an agentic workflow, refining its code by reading frames and analysing motion from the output simulation.}
    }
    \label{fig:method}
\end{figure*}

%% file: sections/method.tex
\section{VDAWorld: VLM-Directed Abstraction and Simulation}
The input to \methodname{} is a pair consisting of a single image of a scene and a text prompt that describes the scene. 
The goal of \methodname{} is to convert this static input into a dynamic, interactive world model. 
This process is orchestrated by a central Vision-Language Model (VLM), which acts as an inverse dynamics model to generate a ``world program'' in Python. 
This generated output captures three key components:
(1) \emph{A Grounded Abstract Representation}: The VLM selects from a suite of vision tools and writes a program to construct a 2D or 3D model of the scene, optimized for simulation,
(2) \emph{Inferred Action Dynamics}: It predicts the most likely abstract action (e.g., an impulse, velocity, or transformation) from the visual and textual cues, which drives the simulation,
(3) \emph{A Simulator Program}: It determines the most compatible simulation method (e.g., rigid body, fluid, logic) to progress the scene's dynamics forward in time.
Once generated, this world program is executed with the inferred action to predict a plausible future. 
Because the system cleanly factors the explicit structured world and its action parameterization, both can be modified with novel user-defined interventions to imagine diverse futures.

\input{sections/formal}

\subsection{Vision API Usage}

An important component of \methodname{} is the provision of a suite of computer vision tools via an API which allow the VLM to build a simulator which matches the provided scene image. We implement a suite of perception and geometry modules that the VLM can call to perform a range of tasks. The VLM does not re-implement or update the implementation of these tools, and it is free to ignore these implementations if it does not find any use for them. 
The toolbox includes perception tools, such as SAM 3~\citep{carion_2025_sam3}; geometry tools such as MoGe-2~\citep{wang2025moge2}; and SAM 3D for mesh extraction from images~\citep{chen_2025_sam3d}.
Primitive fitting methods are also provided to fit geometric shapes such as circles, squares, cubes and cylinders to collections of points using a RANSAC-based approach.
A full specification of the toolbox is provided in supplementary material.

\subsection{Prompting for World Program Generation}
The core of \methodname{} lies in guiding a powerful Vision-Language Model (VLM) to generate a complete, executable world program via an agentic pipeline. 
Instead of fine-tuning, we steer the model's behavior at inference time using a comprehensive, multi-part prompt. 

\paragraph{Task Specification}
The prompt begins with a high-level task specification in natural language. 
This instruction outlines the overall objective: to analyze a user-provided image and text description and produce a self-contained Python script that simulates the scene's future.
It directs the VLM to analyze the user-provided inputs (the image $\mathcal{I}$ and caption $\mathcal{C}$) and produce a self-contained Python script representing both the logical programs $P_s, P_\tau$ as methods, along with the parameterized action $a_{0:T}$.

\paragraph{Environment Specification}
The environment specification provides the formal scaffolding for the VLM's code generation task. 
Its central element is a Python Simulator base class that the VLM must inherit from in its world program. 
This base class defines the core methods the VLM must implement, enforcing the entire simulation logic from scene setup to frame-by-frame execution. 
Crucially, the methods of this class explicitly implement the programs defined in our formal formulation: prescribed methods realize the initial state program $P_s$, the transition program $P_\tau$, and establish the inferred action $a_{0:T}$.
Additionally, the environment provides the VLM with a list of existing python libraries that it can rely on, pointing it to common simulation implementations.

\paragraph{API Specification}
Finally, we provide the VLM with a specification of the diverse computer vision toolkit implemented to support the creation of a simulator which appropriately matches the provided scene image.

\input{figures/qualitative}

\subsection{Agentic Loop}
While a single generation pass can produce high-quality results, high complexity in scenes can lead to errors in the initial code or inaccurate scene fits. 
To enhance the robustness of our system, we utilise an agentic implementation which can iteratively refine and improve its own generations. This is illustrated in Figure~\ref{fig:method}. 
In the agentic implementation, the VLM can independently call a set of tools to interpret its own implementation, and iteratively refine the code it has written.
The agentic tools provided can be broken into three areas.
The first set are implementational tools.
These include the ability to write code from scratch; to edit code with unified diff blocks; to read the documentation associated with each API method; and to execute code.
The second set allow the model to debug its code.
These tools include reading back its own code, reading error messages from the terminal, reading standard output from the terminal, and reading metadata associated with each API call (for example, the number of objects segmented by a single prompt).
Finally, the third set allow the VLM to critique a resulting simulation.
These tools include the ability to view individual frames of simulation.

An important tool we introduce is the ability to view a motion history image~\cite{bobick_2001_mhi}. 
The motion history image visualizes all dynamic activity over the simulation's duration.
The exact formulation of the motion history image we implement uses image hue to encode the time of motion, and image saturation to encode the frequency of motion, while the value channel encodes whether any motion occurred at all. The full equations for this formulation are provided in the supplementary material.
These tools are important for the model's ability to produce coherent accurate simulations (Table \ref{tab:ablation}).

%% file: sections/formal.tex
\subsection{Formal Formulation}
Formally, we consider the task of world modeling as learning a function that maps an initial observation and action conditioning to samples from the distribution over future states. This process is mediated by a generated initial state program $P_s$, a transition program $P_\tau$, and an inferred action sequence $a_{0:T}$. The method can be decomposed into two stages: abstraction and simulation.

\paragraph{Abstraction} The abstraction phase begins with a generative function $\Phi$ which interprets the raw visual inputs and the caption $\mathcal{C}$ to produce a structured world program representation and action sequence:
\begin{equation}
    P_s, a_{0:T}, P_\tau, \mathcal{R} = \Phi(\mathcal{I}, \mathcal{C}).
\end{equation}

All of these components are instantiated as methods and objects within a single executable Python script. Specifically, $P_s$ encapsulates the construction of the scene's representation (geometry, objects, and initial parameters), $a_{0:T}$ encapsulates the predicted action dynamics over time, and $P_\tau$ encapsulates the dynamic simulation engine and update rules.  The last component, $\mathcal{R}$ is a rendering function used to visualize the state of the world at a given timestep.

The action prediction $a_{0:T}$ is jointly inferred alongside the scene representation, meaning the VLM acts not only as a visual parser but formally as an inverse dynamics model or action inference engine. The inferred action sequence $a_{0:T}$ is highly flexible and general in character. 
In physical simulation, $a_t$ might consist of the application of an impulse or a velocity at time $t$.
It might also consist of a geometric transformation of the ego camera, resulting in a simulation where the view of the scene transforms over time.
It could also be the case that the action is essentially null ($a_t = \emptyset$), and the physics of the simulation evolve naturally on account of the starting conditions of the objects in the simulation.

The initial state $s_0$ is given by executing $P_s$ on the image, yielding the structured abstract representation of the scene:
\begin{equation}
    s_0 = \text{Exec}(P_s, \mathcal{I}).
\end{equation}

\paragraph{Simulation} The simulation stage then executes these programs alongside the predicted action to rollout the future states of the world over time $t$. 
Subsequent states are generated by the transition program $P_\tau$ using the state and action at time $t$:
\begin{equation}
    s_{t+1} = \text{Exec}(P_\tau, s_t, a_t),
\end{equation}
where $s_t$ represents the state at time $t$. The scene at any time step can be rendered as $\hat{\mathcal{I}}_t = \mathcal{R}(s_t)$, though we note that this rendered output is strictly apart from the world model state, and is used primarily for visualization and evaluation purposes. The core world model is defined by the state $s_t$ and its dynamics under the transition program $P_\tau$ and action sequence $a_{0:T}$.

The programmatic nature of \methodname{} allows for explicit user intervention, serving as a powerful mechanism for flexible action conditioning. We will demonstrate four primary modes of intervention:
(1) \emph{Caption Modification}: Altering the prompt $\mathcal{C}$ to generate a revised pair of programs and actions.
(2) \emph{Action Modification}: Explicitly editing the code defining the inferred action sequence $a_{0:T}$.
(3) \emph{Transition Function Modification}: Modifying the physical rules or engine logic in $P_\tau$.
(4) \emph{Initial State Modification}: Changing the starting configuration or geometry instantiated by $P_s$.
These enable counterfactual reasoning and control, where the simulation rollout follows the new, action-conditioned dynamics $s'_{t+1} = \text{Exec}(P'_\tau, s'_t, a'_t)$.
By factorizing the world modeling problem through explicit programs $P_s$ and $P_\tau$ and actions $a_{0:T}$, \methodname{} ensures that the generated dynamics are consistent, interpretable, and amenable to structured user intervention.

To practically create the accurate generation of these world states and actions we implement a comprehensive toolbox of vision tools which are available for the VLM to select from.

%% file: figures/qualitative.tex
\begin{figure}[t!]
    \centering
    \includegraphics[width=1.0\linewidth]{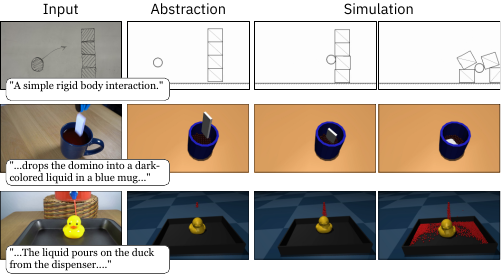}
    \caption{\methodname{} generates an abstraction of the scene and a simulator that can together be used to predict future scene states. From top to bottom: (Abstraction) A collision drawn on a whiteboard, and (Rigid Body) A domino dropping into a mug of liquid, (Fluid Dynamics) Liquid pouring onto a duck in a tray.}
    \label{fig:qualitative}
    \vspace{-0.5cm}
\end{figure}

%% file: sections/results.tex
\section{Experiments}

We evaluate our approach to demonstrate three core advantages over standard video models: (1) flexibility across diverse target domains and applications, (2) physical plausibility via executable code, and (3) interpretability enabling zero-shot control.
Our implementation uses Gemini 3.1 Pro~\citep{deepmind_2025_gemini3} as the core VLM coding agent.
Computations use NVIDIA RTX Pro 6000 Blackwell GPUs. Code will be publicly available.

\subsection{Predicting the Future}
Figure~\ref{fig:qualitative} presents qualitative results of \methodname{} on complex solid body and thermodynamic dynamics. The top row in particular highlights our method's ability to select appropriate abstractions: for a whiteboard drawing, it correctly infers a simple 2D abstraction is sufficient. By intentionally discarding high-frequency visual details, \methodname{} focuses exclusively on producing accurate, physically plausible simulations rather than visual photorealism.

\sig{We use the existing PhysicsIQ benchmark~\cite{physicsiq}, and introduce two new benchmarks: HSPBench, and ConwayBench.}
We benchmark against leading video generation models: Wan2.2 \citep{wan2025} (primary open-source baseline), Lumiere \citep{bar2024lumiere}, VideoPoet \citep{kondratyuk2023videopoet}, and select examples from Veo3~\citep{veo3}\footnote{A full Veo3 evaluation was cost-prohibitive.}. 
We also compare against physical simulation approaches, 
PhysGen~\citep{physgen} and PhysGen3D~\citep{chen2025physgen3d}. As these rely on specific simulators, they represent a subset of \methodname{}'s flexible capabilities.

\subsubsection{PhysicsIQ}
\input{tables/physicsiq}

To demonstrate the flexibility of our approach on standard mechanical reasoning tasks, we evaluate on the PhysicsIQ benchmark~\citep{physicsiq}, which contains real-world videos of diverse physical phenomena. We evaluate on solid mechanics, fluid dynamics, magnetism, and thermodynamics, excluding optics as our method abstracts visual appearance. This results in a total of 174 test samples.
Following the PhysicsIQ protocol, we compute a final score (out of 100) combining Spatial IoU, Weighted Spatial IoU, and Spatiotemporal IoU to evaluate motion accuracy. We omit MSE, as our goal is physically plausible dynamics rather than pixel-perfect reconstruction. \sig{We present results in two settings: a one-trial result as well as a best-of-three-trials result that selects the best performing output among three trials. As the future prediction problem here has inherent uncertainty, we believe the best-of-three setting presents a more accurate picture. }

\begin{figure*}
\centering
\includegraphics[width=0.96\linewidth]{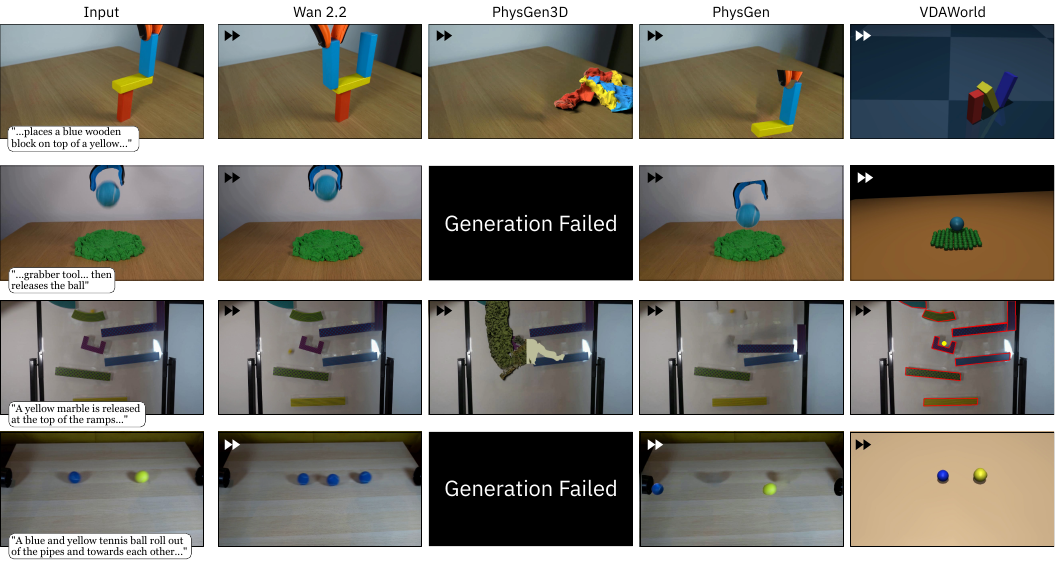}
\caption{Qualitative results on the PhysicsIQ benchmark. \emph{$\triangleright\triangleright$ denotes selecting a frame after iterating the simulation.} Refer to the supplemental webpage for many more results.}
\label{fig:results_qualitative_physicsiq}
\end{figure*}

Table~\ref{tab:physicsiq_results} shows that  
\methodname{} outperforms all baselines in the best-of-three setting, achieving a combined score of 49.7 compared to 46.2 for the strongest video baseline, Wan2.2. In the one-trial setting, VDAWorld slightly outperforms Wan2.2 and significantly outperforms all other baselines. These results show that explicit simulation-based world models can match or exceed pixel-space video models on physical prediction metrics, while additionally providing explicit state, interpretability, and direct counterfactual control.
Our qualitative results (Figure~\ref{fig:results_qualitative_physicsiq} and supplementary video) reveal that the next-best method, Wan2.2, frequently produces non-physical artifacts, such as objects merging, demonstrating the lack of a true underlying physics model, whereas \methodname{} generates physically plausible results.

As PhysGen and PhysGen3D rely on a fixed pipeline with a predetermined simulator, they struggle to generalize across the diverse dynamics in this benchmark, often producing simulations that fail to capture the nuances of true physical behaviour. Notably, this resulted in a high failure rate of $69.9\%$ for PhysGen3D. \sig{\methodname{} is the first demonstration of a simulation-based method that is competitive with state-of-the-art video models.}

\subsubsection{HSPBench}
\input{tables/physics_experiments}
\input{tables/MME-Cof-Pro}

\begin{figure}[h]
    \centering
    \includegraphics[width=\linewidth]{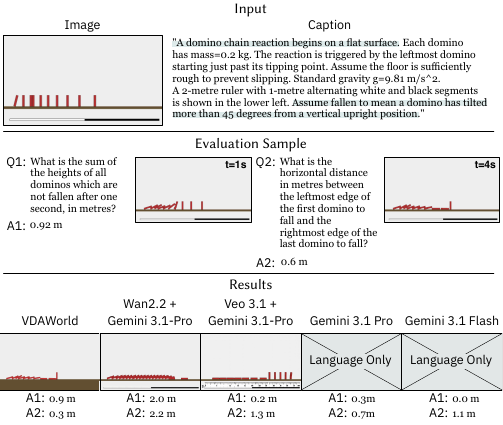}
    \caption{A sample from our High School Physics Benchmark, HSPBench. \methodname{} produces a simulation which reproduces the physics of the original setting more accurately, enabling it to provide high-quality answers to the questions associated with the samples.}
    \label{fig:HSPBench}
\end{figure}

\sig{To explicitly evaluate physical accuracy in a deterministic setting, we present a physics experiment benchmark consisting of 40 high-school level problems (e.g., mass-spring systems), see Figure \ref{fig:HSPBench} for an example.}

Conditioned on an initial state diagram and descriptive caption, these deterministic scenarios cleanly test a model's adherence to physical rules.
\sig{Models are assessed on accuracy compared to objective deviation from these rules using the PhysicsIQ metrics described earlier.} 
A full description of this benchmark is given in supplementary material.
\sig{While video models frequently violate basic physical principles, our approach locks the dynamics into a physics engine, producing significantly better simulations (Table~\ref{tab:physics_experiments} (Bottom) and Figure \ref{fig:HSPBench}). }

\subsubsection{ConwayBench}

\begin{figure}[h]
    \centering
    \includegraphics[width=\linewidth]{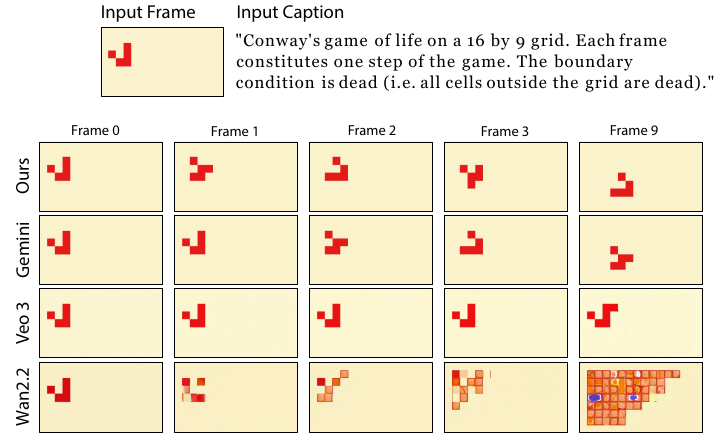}
    \caption{A sample from our Conway's Game of Life benchmark, ConwayBench. Our method is the only one that perfectly captures this logic simulation task. Video models in particular generate significant artifacts.}
    \label{fig:conwayBench}
\end{figure}

Moving beyond physics, we present a benchmark using Conway's Game of Life~\citep{conway1970game}, a discrete cellular automaton. This purely logical setting demonstrates \methodname{}'s flexibility to generate correct simulations in domains governed by abstract rules rather than physical laws. 
\sig{We task the model to predict board evolution from a single input frame across six scenes with different starting configurations and textures, evaluating accuracy per frame via the F1 score.}
Because our approach generates an executable simulation, it acts as a guaranteed correct rule engine, achieving a perfect F1 score and significantly outperforming video models (Figure \ref{fig:kde_conway} (right), Figure~\ref{fig:conwayBench}). 
The Gemini Pro baseline solves simple scenes but fails to parse complex initial states, leading to diverging simulations. State-of-the-art video models are unable to capture the dynamics of this problem and achieve low performance.

\subsubsection{Analysis}

\begin{figure}
\centering
\includegraphics[width=\linewidth]{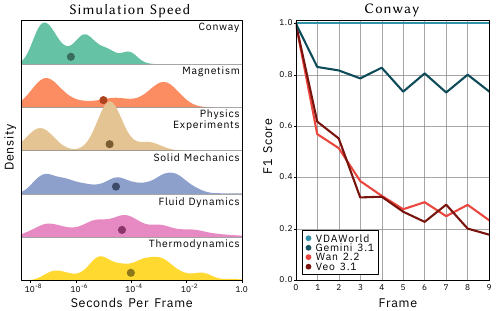}
\caption{\textbf{(Left)} Comparison of simulation execution time densities over all benchmarks using a Kernel Density Estimation plot~\citep{parzen_1962_kde,rosenblatt_1956_kde}. Video models inherently pay a fixed pixel-generation cost, whereas \methodname{} scales with physical complexity, resulting in significantly faster execution for simpler dynamics.  \textbf{(Right)} F1 score computed between ground truth sequences and predicted states on a per-frame basis. On Game of Life, \methodname{} achieves a perfect F1 score, significantly outperforming video models and our Gemini 3.1 Pro baseline. 
}
\label{fig:kde_conway}
\end{figure}

\paragraph{Simulation Time}
We highlight our method's computational efficiency. Unlike video models that incur a large, fixed cost to generate pixels, the execution time (Figure~\ref{fig:kde_conway}, left) of our underlying simulator scales intrinsically with physical complexity. Here, execution time refers specifically to the time taken to run the transition function program $P_\tau$ for simulation.
This aligns with the intuitive requirement of efficient world modeling: simple systems like Game of Life simulate roughly 100$\times$ faster than complex thermodynamics.

Even when considering total computation time, \methodname{} is faster than video models. The total computation time defines the entire inference process.  
Even with the various required stages, on average, computing scenes for the PhysicsIQ benchmark takes 11 minutes for \methodname{}, compared to 30 minutes for Wan2.2.
For the Game of Life benchmark, \methodname{} computes a perfect simulation in 5 minutes on average, while Wan2.2 still takes 30 minutes to generate a diverging video.

\paragraph{Ablation}
\input{tables/ablation}
We ablate the core components of our system on the PhysicsIQ benchmark (best-of-1 setting, central perspective only). 
As shown in Table~\ref{tab:ablation}, all components of our method are essential. 
We also show that using the Gemini Flash model leads to worse performance. 
Finally, we evaluate a VLM-only baseline where Gemini 3.1 Pro is asked to directly generate the output simulation code. This effectively removes API, Tool Use, and MHI all at the same time. 
The VLM-only baseline shows that generic code generation is insufficient. The gains come from the constrained world-program interface, simulation-grounded vision API, action/transition decomposition, and refinement.
These low scores demonstrate that all components of our model are essential.

\subsection{Applications}
\begin{figure}
\centering
\includegraphics[width=\linewidth]{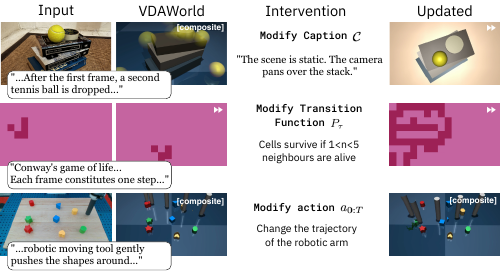}
\caption{Our structured, programmatic representation enables zero-shot user control. \textbf{Row 1:} Caption modification alters dynamics from a collision to a camera pan. \textbf{Row 2:} Transition function modification invents novel Conway rules. \textbf{Row 3:} Action modification provides fine-grained trajectory control over a robotic arm. \emph{[composite] denotes that the image is a composite to visualize the trajectory.}}
\label{fig:results_qualitative_interventions}
\end{figure}

\paragraph{Counterfactual Generation}
Because \methodname{} generates code (the programs $P_s$, $P_\tau$, and action $a_{0:T}$) rather than relying on black-box latents, its representation is fundamentally interpretable. This allows for intuitive, zero-shot human intervention: if physics require adjustment or a counterfactual is desired, a user simply edits the code.
Figure~\ref{fig:results_qualitative_interventions} highlights several intervention modes. \emph{Caption Modification:} Altering the prompt $\mathcal{C}$ changes the simulation from a rigid-body collision to simple camera motion, changing the inferred action sequence. \emph{Transition Function Modification:} Modifying the Game of Life code creates a novel cellular automaton. \emph{Action Modification:} Directly editing a robotic arm's trajectory code provides precise kinematic control, specifying the action conditioning $a'_{0:T}$. Finally, we demonstrate \emph{Initial State Modification} in Figure \ref{fig:teaser}, directly modifying the initial geometry instantiated by $P_s$ to add two more dominoes.

\begin{figure}
    \centering
    \includegraphics[width=\linewidth]{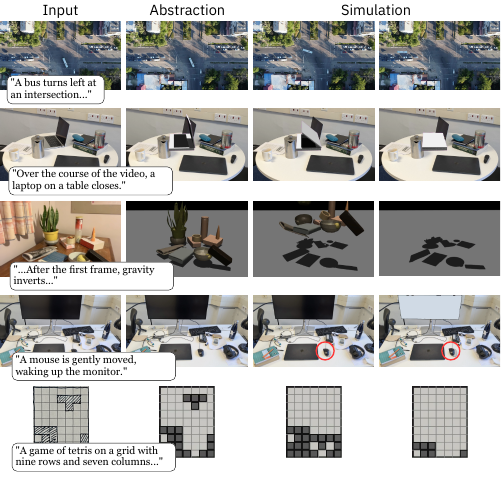}
    \caption{A selection of more complex scenes, highlighting the flexibility and versatility of \methodname{}. These scenes involve aerial traffic scenes, large numbers of interacting objects, and a requirement to abstract a world from a sketch. Note that \methodname{} is required to select the most appropriate level of simulation fidelity to match the complexity of each scene and the desired interaction. Red circle added to emphasise movement.}
    \label{fig:complexity}
\end{figure}

We also demonstrate controllable generation results on some complex and cluttered scenes in Figure \ref{fig:complexity}.

\paragraph{Visual Reasoning}
We first evaluate visual reasoning capabilities of our method using the HSPBench dataset. 
For every scene, we ask two questions about the future that has a deterministically correct answer, e.g., where would the ball in the simulation be on the x-axis after 5 seconds? The input and caption fully specify the initial conditions, making the future evolution deterministic. This gives a total of $80$ question-answer pairs.
We report errors using a VQA error metric that measures the difference between the predicted and GT answer. 
We pass the output simulations of the baselines to Gemini and ask it to interpret the answer. 
We also report the performance of Gemini VLM when it is directly asked for the answer. 
Table~\ref{tab:physics_experiments} demonstrates that we significantly outperform all baselines. 

We also report results on the MME-CoF-Pro benchmark~\cite{qi2025mmecofpro} that includes 303 test samples across 16 categories evaluating capabilities ranging from visual logic to scientific reasoning evaluation.
Example tasks include generating the correct gripper trajectory to place a spatula into a pot and rotating a scene clockwise to make a person upright.
We use their code and metrics (no-hint setting) and report results in Table \ref{tab:mme_comparison}. This benchmark allows us to compare to several contemporary video models, Veo 3.1~\cite{veo3}, Sora2~\cite{sora2}, SeedDance 1.0 \cite{gao2025seedance}, Kling 2.1 \cite{team2025klingavatar}, and Cosmos Transfer \cite{cosmos_transfer}.
\methodname{} outperforms all baselines in this challenging benchmark, demonstrating that code-based simulation can advance visual reasoning significantly.

%% file: tables/physicsiq.tex
\begin{table}[t]
\centering
\resizebox{\linewidth}{!}{
\begin{tabular}{lcccc|c}
\toprule
Model & Solid Mech. & Fluid Dyn. & Thermo. & Magnetism & Combined \\
\midrule
\multicolumn{6}{c}{\textit{Best of three trials}} \\
\midrule
\methodname & \best{54.9} & \second{44.7} & \second{32.6} & \best{25.5} & \best{49.7} \\
PhysGen & 29.7 & 24.6 & 29.6 & \second{23.0} & 28.3 \\
PhysGen3D & 12.2 & 10.2 & 0.4 & 17.2 & 11.7 \\
Wan2.2 & \second{47.3} & \best{49.7} & \best{46.5} & 20.0 & \second{46.2} \\
\midrule
\multicolumn{6}{c}{\textit{One trial}} \\
\midrule
\methodname & \best{42.6} & \second{29.1} & {17.4} & \second{22.0} & \best{37.3} \\
PhysGen & 26.9 & 22.0 & 28.7 & {21.5} & 25.7 \\
PhysGen3D & 6.6 & 7.7 & 0.4 & 17.2 & 7.1 \\
Wan2.2 & \second{39.2} & \best{39.5} & 23.6 & 11.3 & \second{36.8} \\
VideoPoet$^\dagger$ & 35.1 & 24.6 & 29.1 & \best{44.0} & 32.8 \\
Runway & 27.5 & {27.2} & 20.7 & 17.9 & 27.1 \\
Lumiere$^\dagger$ & 27.3 & 23.5 & \best{41.5} & 19.7 & 26.9 \\
Lumiere & 22.0 & 25.4 & \second{33.8} & 19.5 & 23.5 \\
\bottomrule
\end{tabular}
}
\caption{Results on PhysicsIQ. $\dagger$ indicates methods which take videos as inputs. The \colorbox{bestcolor}{best} and \colorbox{secondbestcolor}{second best} numbers are highlighted. \methodname{} outperforms all baselines in both best-of-three and one-trial settings in the final score, and achieves best or second best performance on all categories.
}
\vspace{-0.5cm}
\label{tab:physicsiq_results}
\end{table}

%% file: tables/physics_experiments.tex
\begin{table}[t]
    \centering
    \resizebox{\linewidth}{!}{
    \begin{tabular}{lccc|cc}
    \hline
         & \multicolumn{3}{c}{\textit{Simulation $\rightarrow$ VQA}} & \multicolumn{2}{c}{\textit{VQA}} \\
        \hline
        \textbf{Metric} & VDAWorld & Wan+Gemini & Veo+Gemini & Gemini Pro & Gemini Flash \\
        \hline
        VQA Error $\downarrow$ & \best{75} & 227 & 291 & \second{133} & 175\\
        \hline
        Spatial $\uparrow$ & \best{0.66} & \second{0.19} & 0.22 & - & - \\
        Spatiotemporal $\uparrow$ & \best{0.25} & \second{0.05} & {0.04} & - & - \\
        Weighted Spatial $\uparrow$ & \best{0.73} & {0.28} & \second{0.32} & - & - \\
        Combined $\uparrow$ & \best{0.55} & {0.17} & \second{0.19} & - & - \\
    \end{tabular}
    }
    \caption{Results on HSPBench. \textbf{(Top)} We achieve significantly better scores on VQA on questions about the future state of the scene. \textbf{(Bottom)} We achieve more accurate simulations than the baselines, measured using PhysicsIQ metrics. The \colorbox{bestcolor}{best} and \colorbox{secondbestcolor}{second best} numbers are highlighted. }
    \label{tab:physics_experiments}
    \vspace{-0.5cm}
\end{table}

%% file: tables/MME-Cof-Pro.tex
\begin{table*}[t]
    \centering
    \begin{tabular}{lcccccccc}
        \hline
        {Metric} & Veo3.1 & Veo3.1-fast & Sora2 & Seedance 1.0-pro & Seedance 1.0-fast & Kling-v2.1 & Cosmos Predict & VDAWorld \\
        \hline
        Reasoning Score & \second{55.9} & \second{55.9} & 49.9 & 35.7 & 37.5 & 13.8 & 28.6 & \best{60.5} \\
        Consistency Score & 47.5 & 45.3 & \second{60.5} & 41.5 & 44.5 & 58.3 & 37.5 & \best{74.7} \\
        Average Score & 49.5 & 47.6 & 60.9 & 43.9 & 47.4 & \second{65.1} & 37.8 & \best{73.6} \\
    \end{tabular}
    \caption{Comparison of various methods on the MME-CoF-Pro benchmark. The \colorbox{bestcolor}{best} and \colorbox{secondbestcolor}{second best} numbers are highlighted. \methodname{} outperforms a wide range of state-of-the-art baselines.}
    \label{tab:mme_comparison}
    \vspace{-0.5cm}
\end{table*}

%% file: tables/ablation.tex
\begin{table}[h]
\centering
\begin{tabular}{lcc}
\toprule
Model & Score & Difference \\
\midrule
\methodname & 40.5 & - \\
\midrule
Ablate MHI & 39.0 & -3.8\%\\
Ablate API & 36.2 & -10.7\%\\
Ablate Caption & 35.2 & -13.2\%\\
Ablate Image & 34.6 & -14.6\%\\
Ablate Tool Use & 27.0 & -33.5\%\\
\methodname{} with Gemini 3.1 Flash  & 32.4 & -20.1\%\\
VLM only & 17.7 & -56.4\% \\
\bottomrule
\end{tabular}
\caption{Ablation results demonstrate the significance of the various components of our modelling strategy.}
\vspace{-0.7cm}
\label{tab:ablation}
\end{table}

%% file: sections/limitations.tex
\section{Discussion and Conclusion}
\begin{figure}
    \centering
    \includegraphics[width=\linewidth]{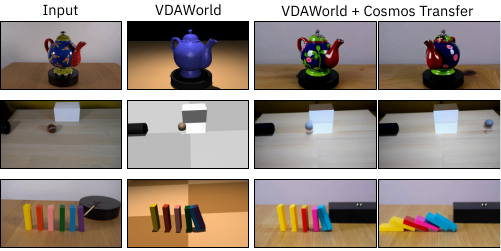}
    \caption{We use Cosmos Transfer \cite{cosmos_transfer} to generate photorealistic results. While the appearance is not perfectly maintained, the results indicate feasibility of this task.}
    \label{fig:cosmos}
\end{figure}

While a central advantage of VDAWorld is that the world model is kept abstract, and while we demonstrate the advantage of this representation in several downstream applications, there may be settings in which reconstructing pixels from simulated videos is desirable. Generating photorealistic videos is not the focus of this paper, and we do not present comprehensive experiments to this effect; however, several off-the-shelf content-transfer methods could potentially address this task. We demonstrate initial results with Cosmos Transfer \cite{cosmos_transfer} on simulations returned by VDAWorld in Figure \ref{fig:cosmos}. Although these are not pixel-perfect reconstructions of the original scene, they suggest that appearance transfer from executable simulations is a promising direction for future work.

VDAWorld is also limited by the quality and coverage of its perception tools, simulator library, and coding VLM. Errors in segmentation or 3D reconstruction can produce simulations that are internally consistent but do not correctly match the input scene. As visual foundation models improve, the fidelity of scenes that VDAWorld can handle will improve as well. 

%% file: sections/conclusion.tex
\paragraph{Conclusion}
In this work we introduced \methodname{}, a new paradigm for building dynamic world models from static images. We have shown that by tasking a Vision-Language Model with world program synthesis, it is possible to generate explicit, executable simulations that are physically plausible, interactive, and versatile. Our experiments demonstrate that this approach avoids common physical artifacts of pixel-prediction models and excels at tasks requiring precise, rule-based reasoning. This programmatic approach represents a significant step towards creating more grounded and interactive world models.
We believe our work points to a broader shift in how we build autonomous agents. Instead of relying on monolithic, end-to-end models that learn an opaque representation of the world, \methodname{} functions as a compositional agent that reasons about the world and writes code to model it.

%% file: sections/acknowledgements.tex
\section{Acknowledgements}

We would like to thank Dr Matthew Johnson for discussions and advice. We acknowledge the use of resources provided by the Isambard-AI National AI Research Resource (AIRR). Isambard-AI is operated by the University of Bristol and is funded by the UK Government’s Department for Science, Innovation and Technology (DSIT) via UK Research and Innovation; and the Science and Technology Facilities Council [ST/AIRR/I-A-I/1023]~\citep{isambard}. We are also grateful to Google Cloud, which provided Gemini credits for use in this project.

%% file: sections/appendix.tex
\section{Appendix}

\subsection{PhysGen3D}

For the PhysGen3D~\cite{chen2025physgen3d} baseline, since no code was provided to derive the physical parameters of a given scene, we simply prompted for the physical parameters.
In our experiments we also imposed a time limit on the generation of a single scene for this baseline to one hour for the timely processing of the Physics IQ benchmark.

\subsection{Ablations}

We complete a thorough ablation of the key components of our pipeline.
These ablations are as follows:

\begin{itemize}
    \item \textbf{Ablate MHI}: We do not allow the agentic VLM to view the motion hisory image; only allowing it to view individual frames from the simulation video.
    \item \textbf{Ablate API}: We do not provide the vision API to the VLM containing the suite of off-the-shelf reconstruction tools.
    \item \textbf{Ablate Caption}: We do not pass the caption to the VLM or to the critic.
    \item \textbf{Ablate Image}: We do not pass the image to the VLM, although it is made available to the method which fits the scene.
    \item \textbf{Ablate Tool Use}: We do give the VLM agentic capabilities, requiring it to simply write the code immediately.
    \item \textbf{Gemini 3.1 Flash}: We use the Gemini 3.1 Flash model rather than 3.1 Pro.
    \item \textbf{VLM} We ablate all features simultaneously, passing only the caption and image to the VLM and requiring it to write a simulation script accepting the image as input.
\end{itemize}

We evaluate these ablations on the central perspective of the Physics IQ dataset.
A table of results is shown in Table~\ref{tab:ablation}.
Results illustrate that the ablation of the motion history has the smallest impact on the PhysicsIQ results.
This is consistent with the fact that this method partially reproduces information available to the VLM via simple frame viewing, and is therefore partially redundant.
On the other hand, ablating either the tool use or using a naïve application of a VLM to this task has a significant impact on the overall results.
This demonstrates the benefits of the structured approach of \methodname over a naive application of a VLM to this world modelling task.

\begin{figure}[h]
    \centering
    \includegraphics[width=\linewidth]{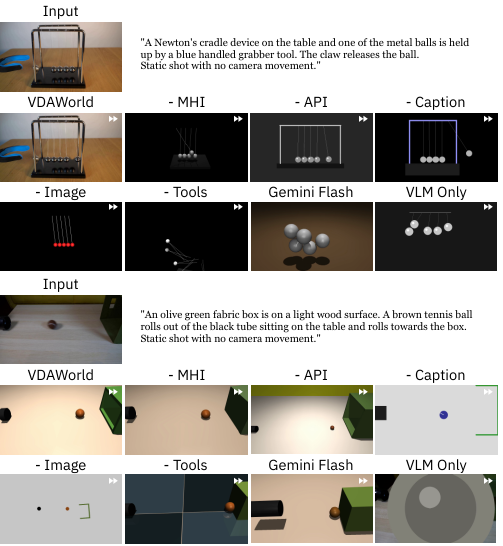}
    \caption{Ablations of VDAWorld. Note that when the image is missing, the appearance of some objects is incorrectly constructed. Without the caption, objects often travel in the wrong direction (for example the ball rolls into the box, rather than behind it). Without the toolbox, objects cannot be accurately matched to their true dimensions. Finally, without the critic, errors in scene construction or dynamics often occur due to coding errors.}
    \label{fig:ablation}
\end{figure}

An illustration of the impact of each ablation is shown in Figure~\ref{fig:ablation}

\subsection{Motion History Image}

The full formula used to calculate the motion history image is as follows:

\begin{equation}
\begin{bmatrix}
\text{hue} \\
\text{sat} \\
\text{val}
\end{bmatrix}_{x,y}
=
\begin{bmatrix}
\frac{1}{2T}\cdot\max \big(t \in [0,T] \text{ where } \hat{\mathcal{I}}_t(x,y) \neq \hat{\mathcal{I}}_{t-1}(x,y)\big) \\
1-\frac{1}{T}\cdot\sum_{t=0}^T \mathbb{I}\big(\hat{\mathcal{I}}_t(x,y) \neq \hat{\mathcal{I}}_{t-1}(x,y)\big)\\
\mathbb{I}\big(\exists t \in [0,T] \text{ where } \hat{\mathcal{I}}_t(x,y) \neq \hat{\mathcal{I}}_{t-1}(x,y)\big)
\end{bmatrix}.
\end{equation}

Where $\mathbb{I}$ is the indicator function for zero, so that $\mathbb{I}(x)=1$ if $x=0$ and $0$ otherwise. 
The image $\hat{\mathcal{I}}_t$ is the rendered image of the world model at timestep $t$. 

\begin{figure}
    \centering
    \includegraphics[width=\linewidth]{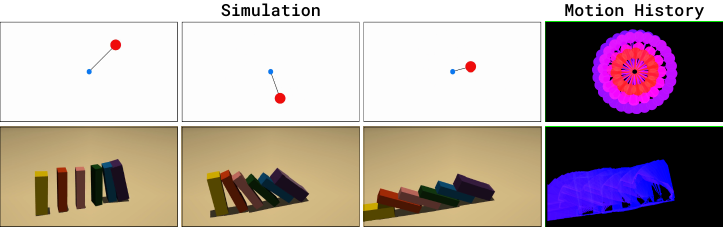}
    \caption{Example motion history images for two videos. These are made available to the model for refinement.}
    \label{fig:motion_history}
\end{figure}

The objective of this formulation is to provide maximum information to the critic about exactly what motion occurs within each video, and where this motion occurs.
Two examples of motion history images are shown in Figure~\ref{fig:motion_history}.

\subsection{Conway's Game of Life Benchmark}

In this section we describe the design of Conway's Game of Life benchmark.

\begin{figure}
    \centering
    \includegraphics[width=\linewidth]{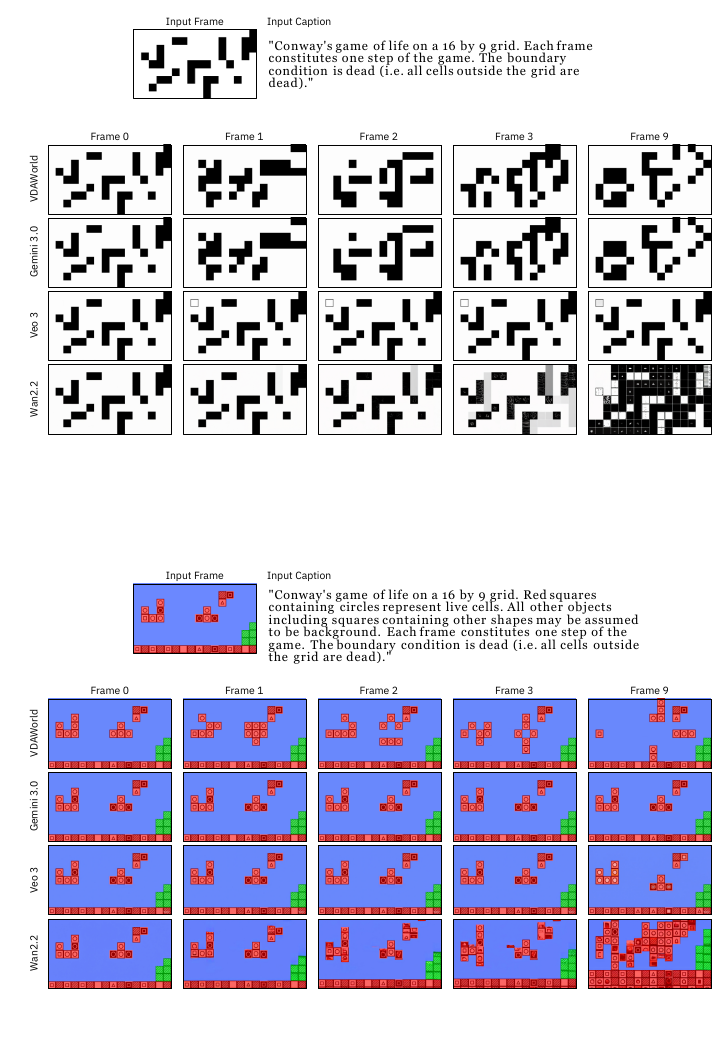}
    \caption{Conway's Game of Life results 1.}
    \label{fig:conway_0}
\end{figure}

\begin{figure}
    \centering
    \includegraphics[width=\linewidth]{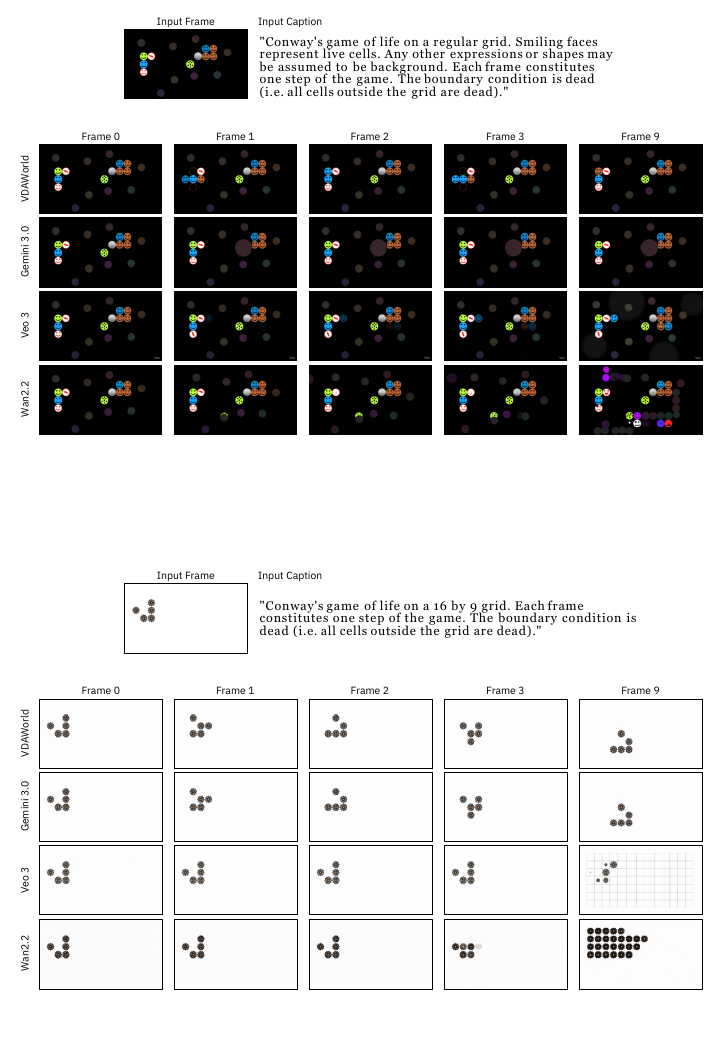}
    \caption{Conway's Game of Life results 2.}
    \label{fig:conway_1}
\end{figure}

\begin{figure}
    \centering
    \includegraphics[width=\linewidth]{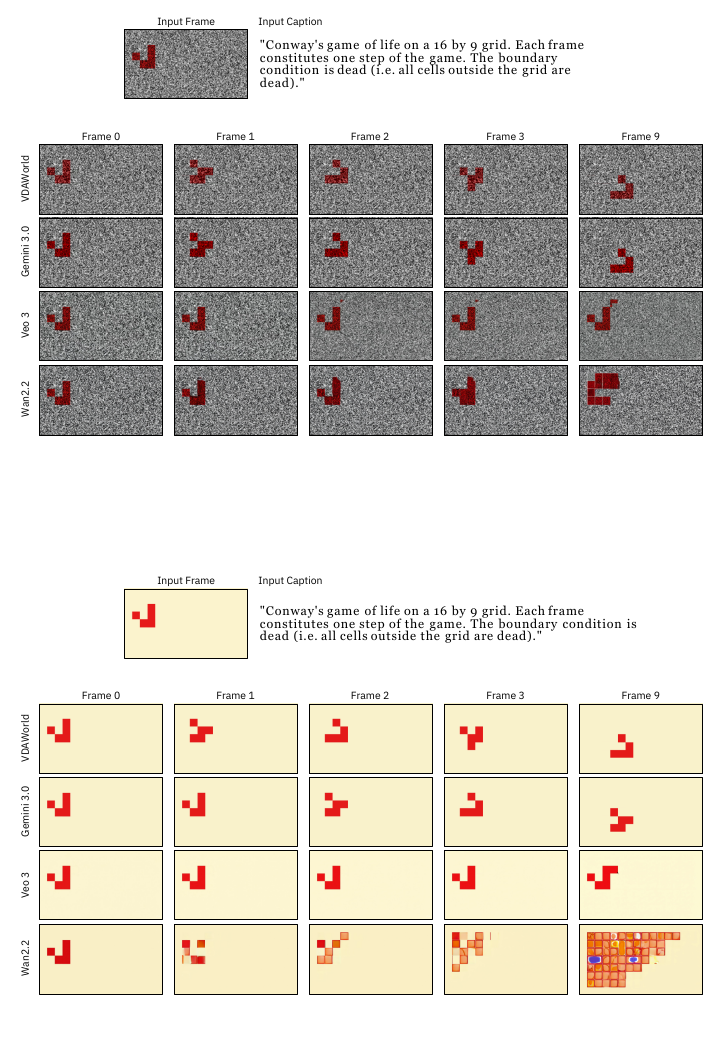}
    \caption{Conway's Game of Life results 3.}
    \label{fig:conway_2}
\end{figure}

Conway's Game of Life~\cite{conway1970game} is a deterministic, rule based game.
This makes it a compelling setting for evaluation of strict rule adherence by various world modelling style methods.
We produce six different image/caption pairs illustrating a starting state of the Game, and providing instructions about the setup and conditions.
Each of the pairs of image/captions, and results from each approach evaluated, are shown in Figures~\ref{fig:conway_0},~\ref{fig:conway_1}, and~\ref{fig:conway_2}.

We additionally show a breakdown of results in Table~\ref{tab:conway_results}.
Results demonstrate the advantage of our approach, systematically producing correct simulations of each scenario.

\begin{table}[h]
\centering
\begin{tabular}{lcccc}
\toprule
Frame & \methodname & Gemini 3.0 & Veo 3 & Wan2.2 \\
\midrule
0 & 1.000 & 1.000 & 1.000 & 1.000 \\
1 & 1.000 & 0.830 & 0.617 & 0.569 \\
2 & 1.000 & 0.817 & 0.552 & 0.514 \\
3 & 1.000 & 0.786 & 0.323 & 0.386 \\
4 & 1.000 & 0.827 & 0.325 & 0.329 \\
5 & 1.000 & 0.734 & 0.267 & 0.277 \\
6 & 1.000 & 0.805 & 0.228 & 0.304 \\
7 & 1.000 & 0.731 & 0.295 & 0.250 \\
8 & 1.000 & 0.800 & 0.201 & 0.294 \\
9 & 1.000 & 0.735 & 0.178 & 0.234 \\
\bottomrule
\end{tabular}
\caption{Full results on Conway's Game of Life benchmark. F1 scores for video methods drop quickly as a result of diverging simulations and failure to adhere to rules. While Gemini 3.0 is able to model simpler Conway settings, it cannot identify challenging starting arrangements. By contrast, \methodname produces perfect simulations in each instance.}
\label{tab:conway_results}
\end{table}

\subsection{Physics Experiments Benchmark}

We contribute the Physics Experiments Benchmark, a set of 40 high school-level mechanics problems for solving by a world model.
The 40 problems are formed by taking 4 numerical variations of ten problem settings.
Each setting is deterministic both due to the setup of the environment and due to information provided in the caption, so the outcome should be fixed in each case.
For each video, the world model must provide a response to two questions. Both questions relate to the measurement of some factor (a distance travelled, a distance between two objects, etc.) in metres.
For consistency between scenes at different scales, for evaluation we convert these measurements back into pixel space.

We report four different scores as part of HSPBench. One is an accuracy over the VQA task. The remaining three are taken from the evaluation procedure of the PhysicsIQ benchmark.

\begin{itemize}
\item \textbf{VQAError} measures the simple mean absolute error between the response to each question in the dataset.
\item \textbf{Spatial IoU} measures the IoU between binary masks recording any pixel which changed over the course of the video between the ground truth simulation and the simulation provided by the model, capturing an impression of where action happens in the simulation.
\item \textbf{Spatiotemporal IoU} measures the IoU between binary tensor masks recording each pixel which changes at each recorded frame, capturing the model's understanding of where and when motion occurs in each simulation.
\item \textbf{Weighted Spatial IoU} measures the IoU between the mean of those binary tensors used in the computation of Spatiotemporal IoU along the temporal dimension according to $IoU=2\frac{\sqrt{\textrm{GT}\cdot\textrm{Pred}}}{\textrm{GT}+\textrm{Pred}}$. This captures the model's understanding of how much motion occurs at each pixel in each frame of the simulation.
\end{itemize} 

\begin{figure}
    \centering
    \includegraphics[width=\linewidth]{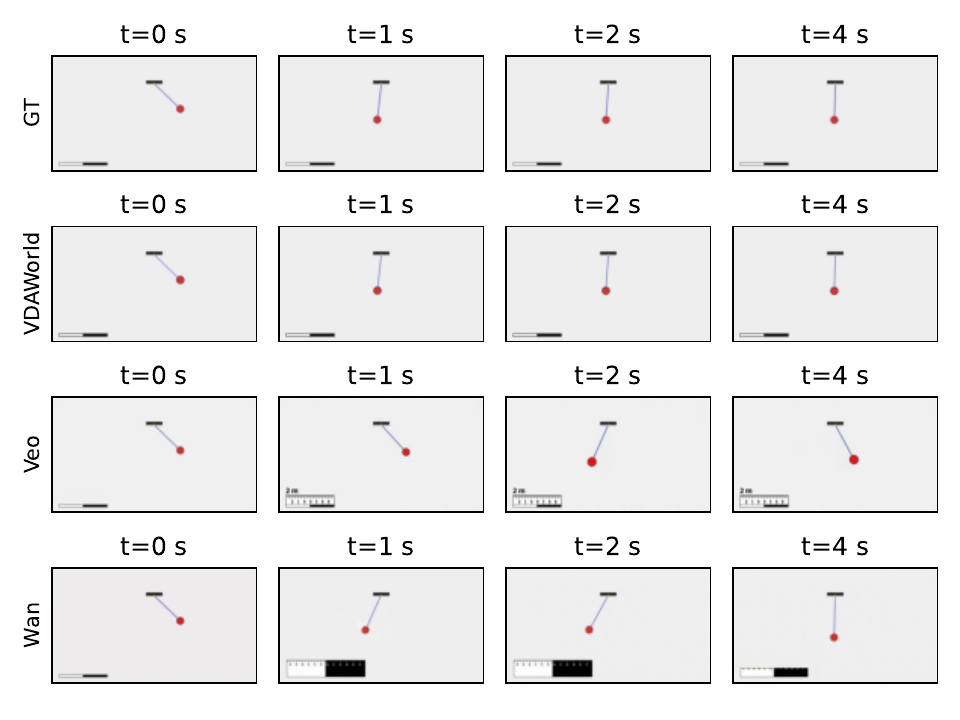}
    \caption{HSPS dataset entry with description: A damped pendulum of bob mass m=0.5 kg and rotational pivot damping $c=2.0 kg*m^2/s$ is released from rest. Standard gravity $g=9.81 m/s^2$ acts downwards. A 2-meter ruler with 1-meter alternating white and black segments is shown in the lower left.}
    \label{fig:exp1_run1}
\end{figure}

\begin{figure}
    \centering
    \includegraphics[width=\linewidth]{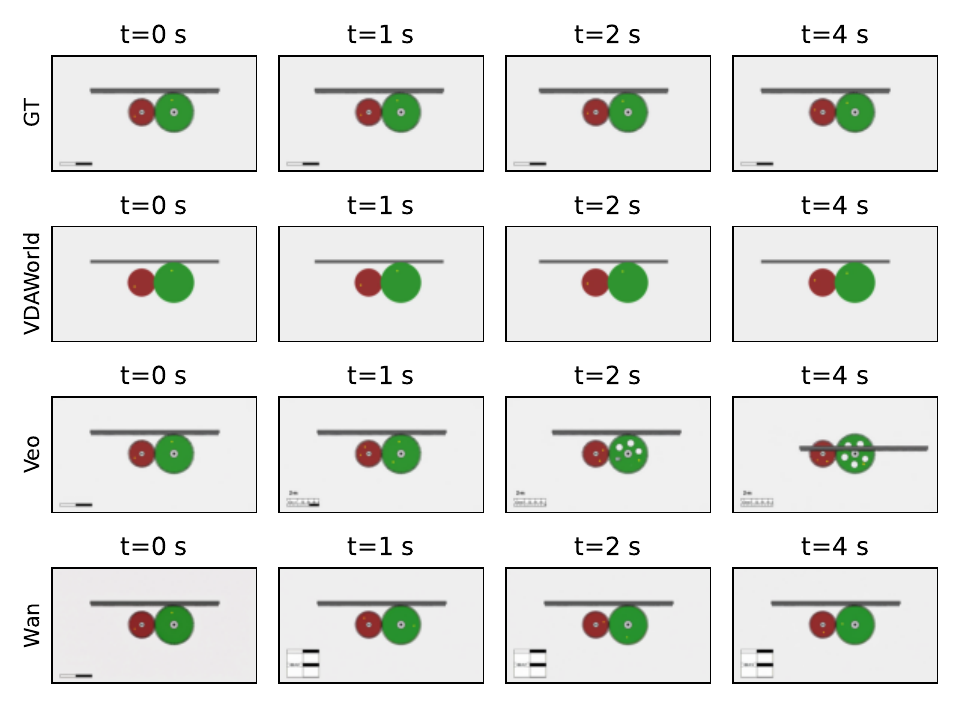}
    \caption{HSPS dataset entry with description: A rack is connected to a gear train. The red gear drives the system, rotating at constant angular velocity 0.2 rad/s clockwise. The green driven gear is meshed with the rack. A 2-metre ruler with 1-metre alternating white and black segments is shown in the lower left.}
    \label{fig:exp3_run1}
\end{figure}

\begin{figure}
    \centering
    \includegraphics[width=\linewidth]{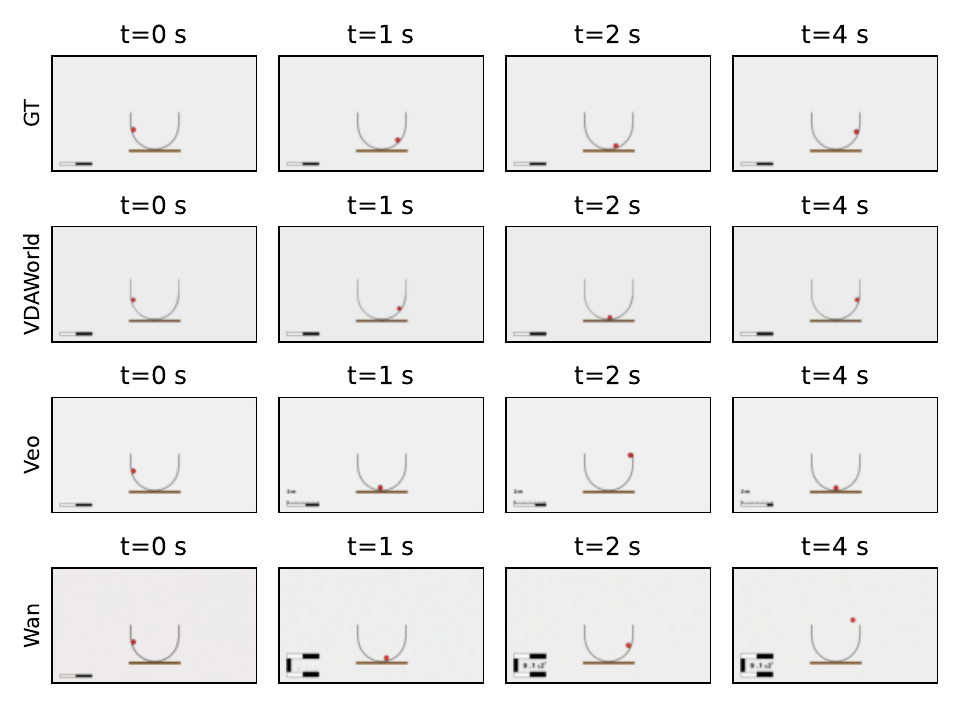}
    \caption{HSPS dataset entry with description: A ball is dropped from rest into a frictionless half-pipe with circular cross-section radius R=1.5 m. The ball is of mass m=0.5 kg. Standard gravity $g=9.81 m/s^2$ acts downwards. No friction acts anywhere. A 2-metre ruler with 1-metre alternating white and black segments is shown in the lower left.}
    \label{fig:exp5_run1}
\end{figure}

\begin{figure}
    \centering
    \includegraphics[width=\linewidth]{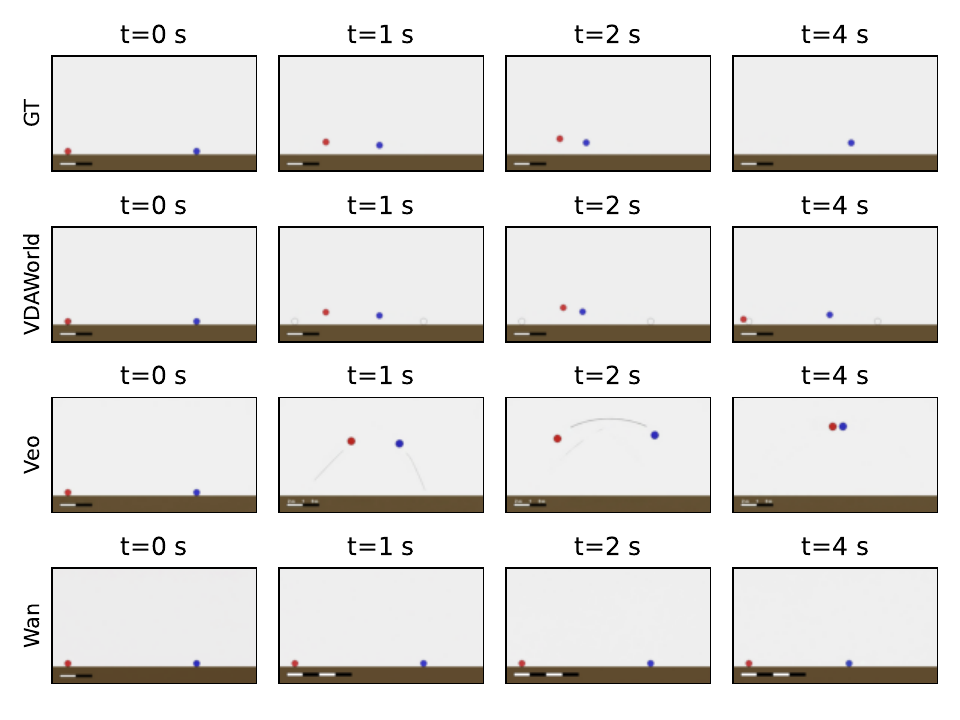}
    \caption{HSPS dataset entry with description: A red ball is launched from ground level at 4.0 m/s at an angle of 60 degrees above the horizontal and to the right. A second blue ball starts from x=8.0 m on the right and is launched at 45 degrees above the horizontal to the left, with the same coefficient of restitution as the floor and the ball-ball collision. The floor has a coefficient of restitution of 1.0 and is frictionless. Standard gravity $g=9.81 m/s^2$ acts downwards. A 2-metre ruler with 1-metre alternating white and black segments is shown in the lower left.}
    \label{fig:exp6_run1}
\end{figure}

\begin{figure}
    \centering
    \includegraphics[width=\linewidth]{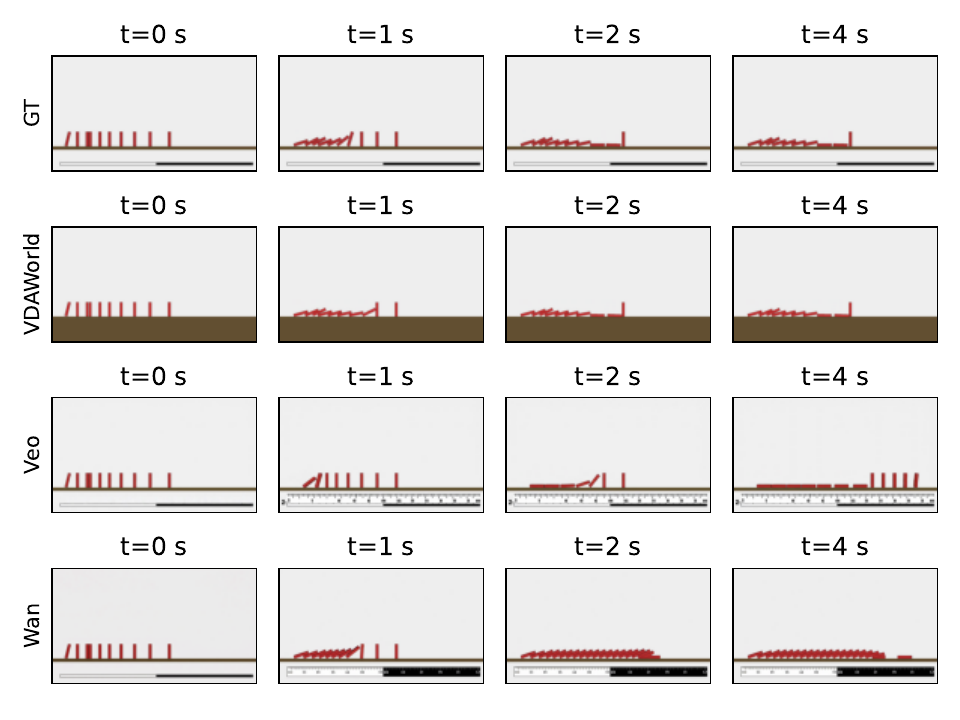}
    \caption{HSPS dataset entry with description: A domino chain reaction begins on a flat surface. Each domino has mass=0.05 kg. The reaction is triggered by the leftmost domino starting just past its tipping point. Assume the floor is sufficiently rough to prevent slipping. Standard gravity $g=9.81 m/s^2$. A 2-metre ruler with 1-metre alternating white and black segments is shown in the lower left. Assume fallen to mean a domino has tilted more than 45 degrees from a vertical upright position.}
    \label{fig:exp8_run1}
\end{figure}

\begin{figure}
    \centering
    \includegraphics[width=\linewidth]{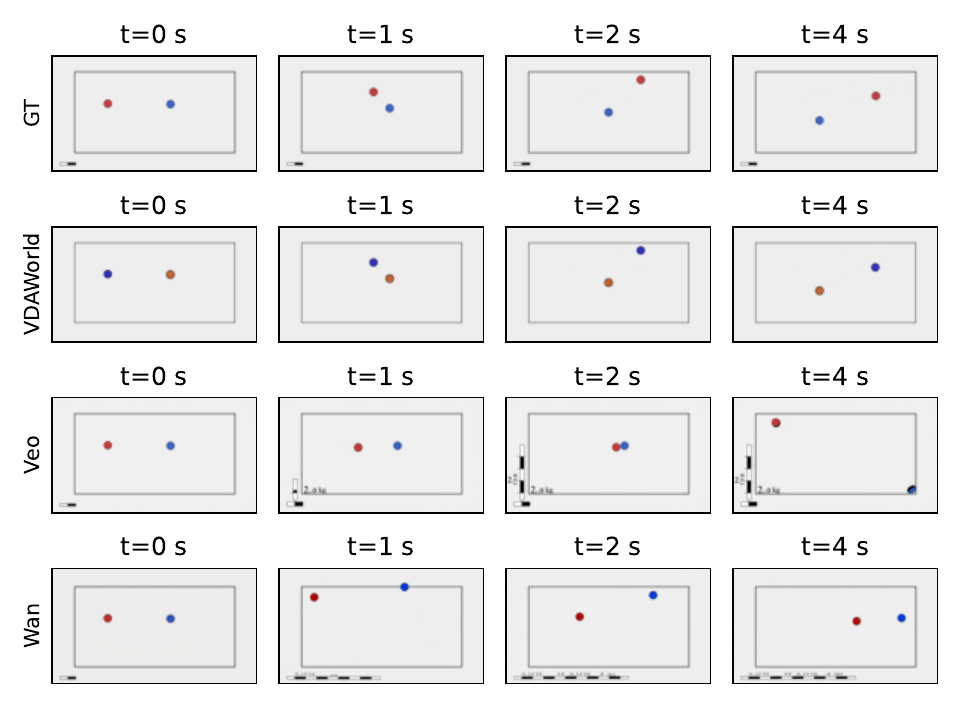}
    \caption{HSPS dataset entry with description: Two circular pucks move on a frictionless surface in a rectangular bounding box. The red puck (mass 2.0 kg) starts with velocity (5.0, 1.5) m/s. The blue puck (mass 1.0 kg) starts with velocity (-1.0, -0.5) m/s. All collisions are perfectly elastic and frictionless. A 2-metre ruler with 1-metre alternating white and black segments is shown in the lower left. The coordinate system is defined with positive x to the right and positive y upwards.}
    \label{fig:exp11_run1}
\end{figure}

\begin{figure}
    \centering
    \includegraphics[width=\linewidth]{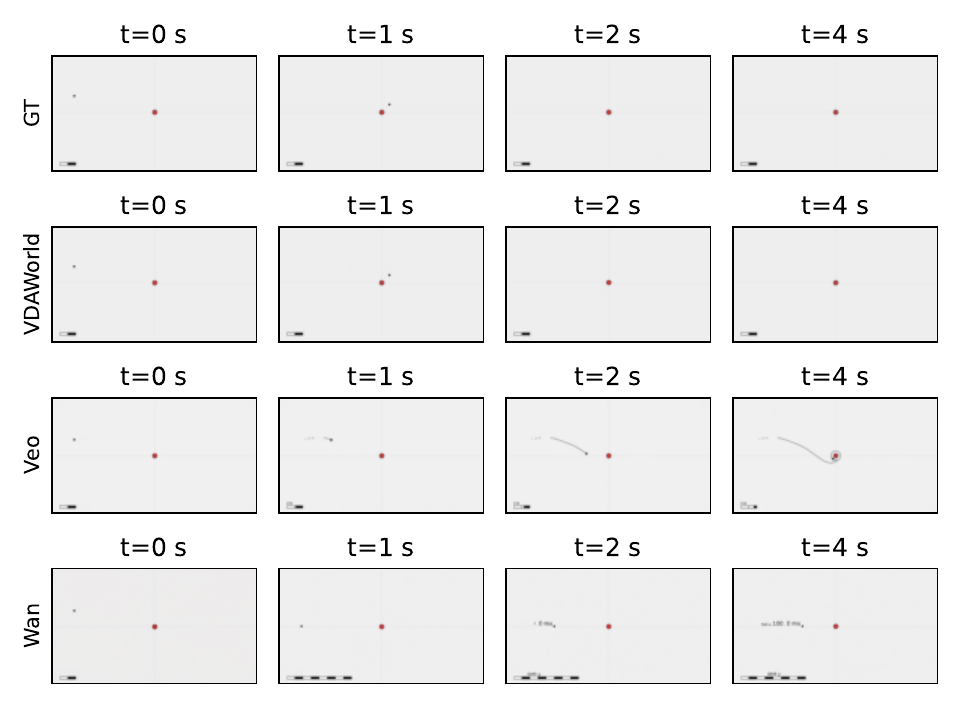}
    \caption{HSPS dataset entry with description: A small black mass begins the simulation with velocity only in the x direction of (vx=10.0m/s). The central red body has gravitational parameter GM=100 m³/s², where GM is the product of the gravitational constant G and the central body's mass — the quantity that determines gravitational acceleration. Assume there is no drag on the system. A 2-metre ruler with 1-metre alternating white and black segments is shown in the lower left. The coordinate system is such that positive x is to the right. }
    \label{fig:exp13_run1}
\end{figure}

\begin{figure}
    \centering
    \includegraphics[width=\linewidth]{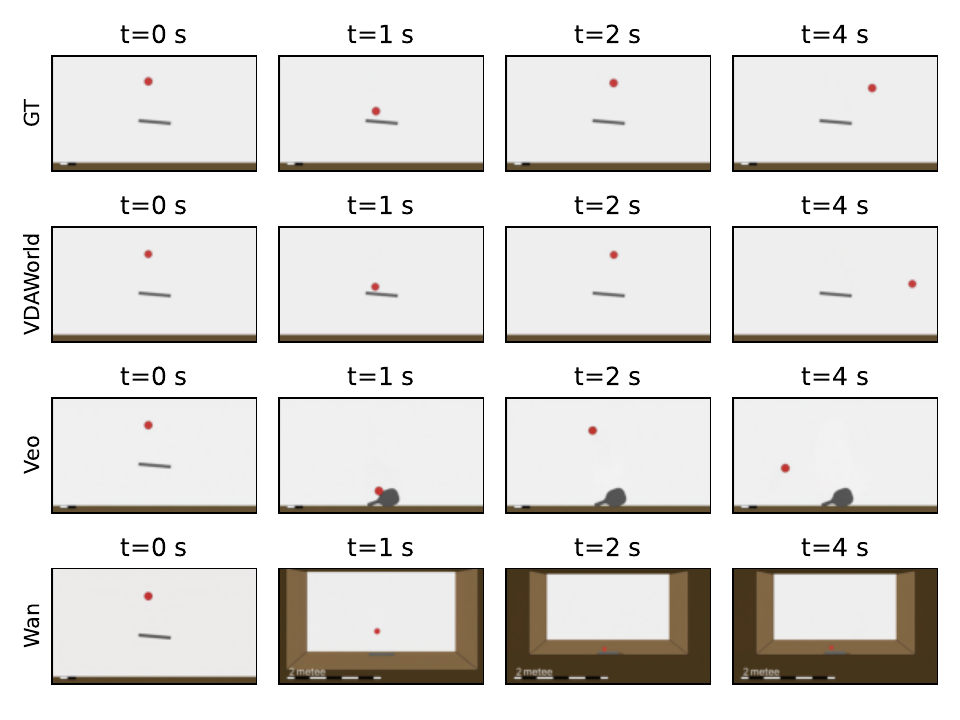}
    \caption{HSPS dataset entry with description: A ball is dropped from rest in an environment contained on all sides by walls. It falls towards a rigid paddle. The paddle, walls and floor are all perfectly elastic and frictionless. A 2-metre ruler with 1-metre alternating white and black segments is shown in the lower left.}
    \label{fig:exp15_run0}
\end{figure}

\begin{figure}
    \centering
    \includegraphics[width=\linewidth]{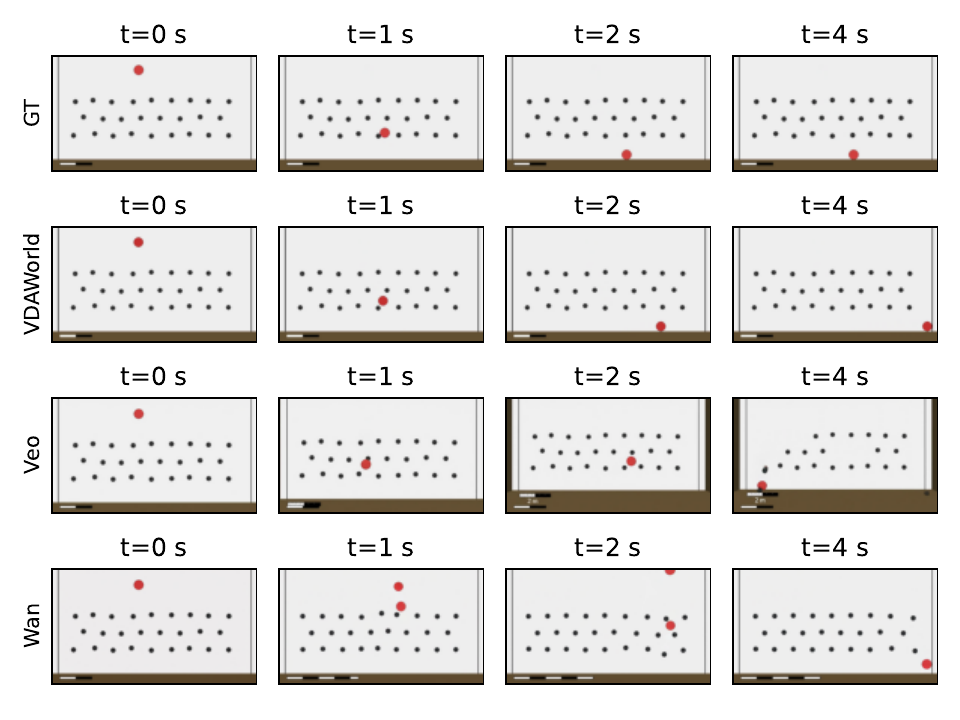}
    \caption{HSPS dataset entry with description: A red ball is released from rest from x=-1.0 m above a deterministic pin board. The ground is bounded by walls on either side and is totally frictional (friction coefficient 1.0 for the ground and the ball), so the ball stops as soon as it lands. There is zero friction between the ball and the pins and the walls. The coefficient of restitution is zero for all interactions. A 2-metre ruler with 1-metre alternating white and black segments is shown in the lower left.}
    \label{fig:exp16_run1}
\end{figure}

An example from each of the ten problem settings is as follows:
\begin{enumerate}
\item \textbf{Damped Pendulum}, Figure~\ref{fig:exp1_run1}: A damped pendulum of bob mass $m=0.5 kg$ and rotational pivot damping $c=2.0 kg*m^2/s$ is released from rest. Standard gravity $g=9.81 m/s^2$ acts downwards. A 2-meter ruler with 1-meter alternating white and black segments is shown in the lower left.

\textbf{Q1}: After 1s, what is the magnitude of the x-displacement of the red mass relative to its pivot (in metres)?\textbf{A1}: $0.20$

\textbf{Q2}: What is the total distance travelled by the mass between t=1 and t=4 s (in metres)?\textbf{A2}: $0.78$

\item \textbf{Rack and Pinion}, Figure~\ref{fig:exp3_run1}: A rack is connected to a gear train. The red gear drives the system, rotating at constant angular velocity $0.2 rad/s$ clockwise. The green driven gear is meshed with the rack. A 2-metre ruler with 1-metre alternating white and black segments is shown in the lower left.

\textbf{Q1}: What is the magnitude of the x-displacement of the left edge of the rack relative to the central pivot of the red gear after $4s$ in metres?\textbf{A1}: $3.84$

\textbf{Q2}: What is the distance between the two yellow dots fixed to each gear after $4s$ in metres?\textbf{A2}: $2.05$

\item \textbf{Ball in Half Pipe}, Figure~\ref{fig:exp5_run1}: A ball is dropped from rest into a frictionless half-pipe with circular cross-section radius $R=1.5 m$. The ball is of mass $m=0.5 kg$. Standard gravity $g=9.81 m/s^2$ acts downwards. No friction acts anywhere. A 2-metre ruler with 1-metre alternating white and black segments is shown in the lower left.

\textbf{Q1}: How many metres does the ball travel in the first two seconds?\textbf{A1}: $5.81$

\textbf{Q2}: What is the height of the ball above the base plate in metres after 3s?\textbf{A2}: 1.00
\item \textbf{Two Ball Bouncing}, Figure~\ref{fig:exp6_run1}: A red ball is launched from ground level at $4.0 m/s$ at an angle of $60$ degrees above the horizontal and to the right. A second blue ball starts from $x=8.0 m$ on the right and is launched at $45$ degrees above the horizontal to the left, with the same coefficient of restitution as the floor and the ball-ball collision. The floor has a coefficient of restitution of $1.0$ and is frictionless. Standard gravity $g=9.81 m/s^2$ acts downwards. A 2-metre ruler with 1-metre alternating white and black segments is shown in the lower left.

\textbf{Q1}: What is the height of the red ball in metres at $t=2s$?\textbf{A1}: $0.92$

\textbf{Q2}: What is the distance between the balls in metres at t=3s?\textbf{A2}: $4.94$
\item \textbf{Domino Chain}, Figure~\ref{fig:exp8_run1}: A domino chain reaction begins on a flat surface. Each domino has $mass=0.05 kg$. The reaction is triggered by the leftmost domino starting just past its tipping point. Assume the floor is sufficiently rough to prevent slipping. Standard gravity $g=9.81 m/s^2$. A 2-metre ruler with 1-metre alternating white and black segments is shown in the lower left. Assume fallen to mean a domino has tilted more than $45$ degrees from a vertical upright position.

\textbf{Q1}: What is the horizontal distance in metres between the leftmost edge of the first domino to fall and the rightmost edge of the last domino to fall?\textbf{A1}: $0.92$

\textbf{Q2}: What is the sum of the heights of all dominos which are not fallen after one second, in metres?\textbf{A2}: $0.60$
\item \textbf{Two Pucks}, Figure~\ref{fig:exp11_run1}: Two circular pucks move on a frictionless surface in a rectangular bounding box. The red puck (mass 2.0 kg) starts with velocity (5.0, 1.5) m/s. The blue puck (mass 1.0 kg) starts with velocity (-1.0, -0.5) m/s. All collisions are perfectly elastic and frictionless. A 2-metre ruler with 1-metre alternating white and black segments is shown in the lower left. The coordinate system is defined with positive x to the right and positive y upwards.

\textbf{Q1}: After 4s, what is the distance between the red and blue pucks (in metres)?\textbf{A1}: $7.62$

\textbf{Q2}: Within the first four seconds, at the moment when the blue puck reaches its lowest position, what is the horizontal distance between the red puck and the left wall (in metres)?\textbf{A2}: $15.00$
\item \textbf{Orbital Motion}, Figure~\ref{fig:exp13_run1}: A small black mass begins the simulation with velocity only in the x direction of ($vx=10.0m/s$). The central red body has gravitational parameter $GM=100 m^3/s^2$, where GM is the product of the gravitational constant G and the central body's mass — the quantity that determines gravitational acceleration. Assume there is no drag on the system. A 2-metre ruler with 1-metre alternating white and black segments is shown in the lower left. The coordinate system is such that positive x is to the right. 

\textbf{Q1}: After two seconds, what is the distance between the black planet and the left edge of the screen in metres?\textbf{A1}: $20.55$

\textbf{Q2}: How far does the black mass travel in metres between t=0s and t=2s?\textbf{A2}: $22.62$
\item \textbf{Ball and Paddle}, Figure~\ref{fig:exp15_run0}: A ball is dropped from rest in an environment contained on all sides by walls. It falls towards a rigid paddle. The paddle, walls and floor are all perfectly elastic and frictionless. A 2-metre ruler with 1-metre alternating white and black segments is shown in the lower left.

\textbf{Q1}: What is the distance in metres between the ball and the lower left corner of the frame after t=4 seconds?\textbf{A1}: $19.61$

\textbf{Q2}: What is the distance travelled by the ball in metres in the x-direction between t=1 and t=4 seconds?\textbf{A2}: $5.24$
\item \textbf{Galton Board}, Figure~\ref{fig:exp16_run1}: A red ball is released from rest from $x=-1.0 m$ above a deterministic pin board. The ground is bounded by walls on either side and is totally frictional (friction coefficient $1.0$ for the ground and the ball), so the ball stops as soon as it lands. There is zero friction between the ball and the pins and the walls. The coefficient of restitution is zero for all interactions. A 2-metre ruler with 1-metre alternating white and black segments is shown in the lower left.

\textbf{Q1}: What is the distance in metres between the ball's starting position and its position after 4s?\textbf{A1}: $5.63$

\textbf{Q2}: What is the largest distance in metres between any two pins the ball collides with in the first 4s? If only one pin is collided with, return 0.0m.\textbf{A2}: $2.38$
\item \textbf{Reflection/Absorption}, Figure~\ref{fig:exp21_run1}: A laser pointer emits a red beam that travels at 2.0 metres per second horizontally and to the right. The beam reflects off grey paddles and is absorbed by black paddles. The laser pointer's origin is indicated with a blue marker. A 2-metre ruler with 1-metre alternating white and black segments is shown in the lower left.

\textbf{Q1}: How many metres does the light travel between the first and third seconds?\textbf{A1}: $4.00$

\textbf{Q2}: What is the distance between the centroid of the first paddle the light is incident on, and the last paddle the light is incident on within 4s?\textbf{A2}: $2.42$
\end{enumerate}

\clearpage
\begin{figure}
    \centering
    \includegraphics[width=\linewidth]{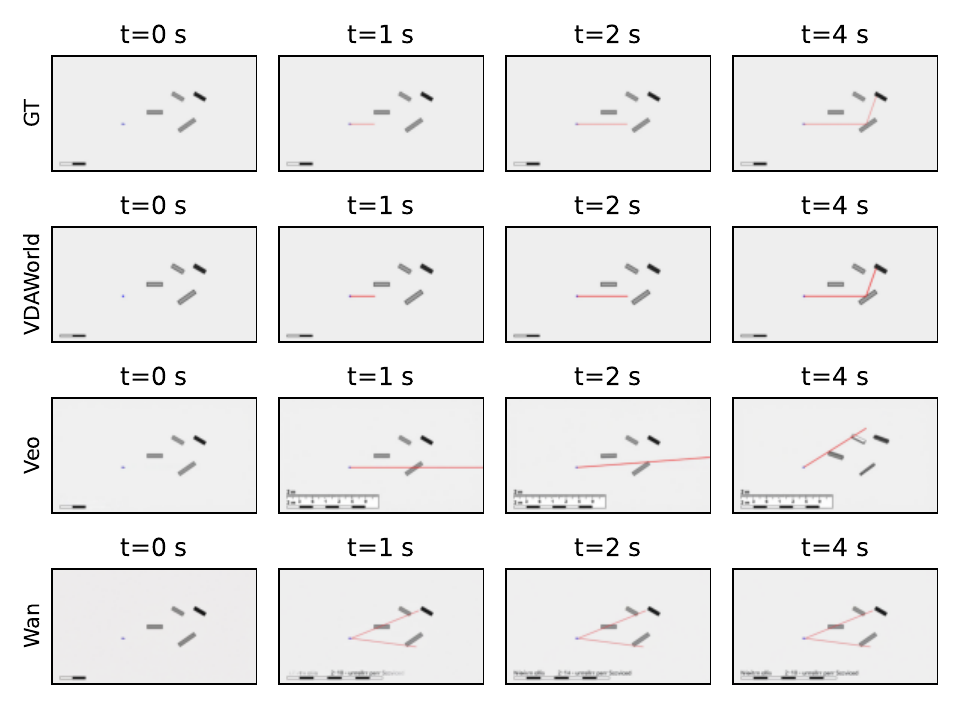}
    \caption{HSPS dataset entry with description: A laser pointer emits a red beam that travels at 2.0 metres per second horizontally and to the right. The beam reflects off grey paddles and is absorbed by black paddles. The laser pointer's origin is indicated with a blue marker. A 2-metre ruler with 1-metre alternating white and black segments is shown in the lower left.}
    \label{fig:exp21_run1}
\end{figure}